\documentclass[10pt,twocolumn,letterpaper]{article}

\usepackage{cvpr}              

\usepackage{kotex}
\usepackage{tikz}
\usepackage{comment}
\usepackage{color}
\usepackage[accsupp]{axessibility}

\usepackage{enumitem}
\usepackage{graphicx}
\usepackage{amsmath}
\usepackage{booktabs}
\usepackage{mathtools}
\usepackage{url}
\usepackage{comment}
\usepackage{color}
\usepackage{graphicx,setspace}
\usepackage{multirow, array, booktabs,makecell,xspace,bm}
\usepackage{pifont}
\usepackage{amsfonts}
\usepackage{arydshln}
\usepackage{dsfont}

\usepackage{stackengine}

\usepackage{scalerel}
\usepackage{esdiff} 

\usepackage{bbm}

\usepackage{algorithm}
\usepackage{algpseudocode}

\usepackage[most]{tcolorbox}

\algnewcommand{\LineComment}[1]{\State \# #1} 

\newcommand{\cmark}{\ding{51}}%
\definecolor{cvprblue}{rgb}{0.21,0.49,0.74}
\usepackage[pagebackref,breaklinks,colorlinks,allcolors=cvprblue]{hyperref}
\def\paperID{698} 
\def\confName{CVPR}
\def\confYear{2025}

\title{Object-aware Sound Source Localization via Audio-Visual Scene Understanding}

\author{Sung Jin Um$^{1,}$\footnotemark[1] \qquad
Dongjin Kim$^{1,2,}$\thanks{Equal contribution} \qquad
Sangmin Lee$^{3,}$\thanks{Corresponding author} \qquad
Jung Uk Kim$^{1,}$\footnotemark[2]\\
$^{1}$Kyung Hee University \qquad
$^{2}$KAIST AI \qquad
$^{3}$Sungkyunkwan University\\
{\tt\small \{sungzin1, ju.kim\}@khu.ac.kr, dj\_kim@kaist.ac.kr, sangmin.lee@skku.edu}
}
\newtcolorbox{mybox2}[1][]{%
    enhanced,
    top=1pt,
    bottom=1pt,
    boxrule=0pt,
    arc=0pt,
    colframe=white,
    colback=white,
    borderline north={0.5mm}{0mm}{black},
    borderline south={0.5mm}{0mm}{black},
    width=\textwidth, 
    left=0pt,
    right=0pt,
    #1
}
\begin{document}
\maketitle
\begin{abstract}
Audio-visual sound source localization task aims to spatially localize sound-making objects within visual scenes by integrating visual and audio cues. However, existing methods struggle with accurately localizing sound-making objects in complex scenes, particularly when visually similar silent objects coexist. This limitation arises primarily from their reliance on simple audio-visual correspondence, which does not capture fine-grained semantic differences between sound-making and silent objects. To address these challenges, we propose a novel sound source localization framework leveraging Multimodal Large Language Models (MLLMs) to generate detailed contextual information that explicitly distinguishes between sound-making foreground objects and silent background objects. To effectively integrate this detailed information, we introduce two novel loss functions: Object-aware Contrastive Alignment (OCA) loss and Object Region Isolation (ORI) loss. Extensive experimental results on MUSIC and VGGSound datasets demonstrate the effectiveness of our approach, significantly outperforming existing methods in both single-source and multi-source localization scenarios. Code and generated detailed contextual information are available at: \href{https://github.com/VisualAIKHU/OA-SSL}{\color{magenta}{https://github.com/VisualAIKHU/OA-SSL}}.
\end{abstract}    

\section{Introduction}
\label{sec:intro}
In recent years, there has been a growing emphasis on audio-visual multimodal tasks in computer vision \cite{liu2024event_detection, ivanko2023speech_recognition, owens2018scene_analysis, ruan2023video_generation, s_CVPR2018_Senocak, um2025watch}. These tasks aim to solve complex challenges by combining auditory and visual information, enabling models to interpret scenes and capture both the spatial and temporal dynamics of real-world environments.

Among these various multimodal tasks, audio-visual sound source localization has been receiving significant attention \cite{s_CVPR2018_Senocak, s_CVPR2021_lvs, s_hard_positive_mining, s_iterative2023, s_um2023sira, s_sun2023learning, s_WACV2022_Shi, s_WACV2023_htf,s_momentum,s_flowgrad, s_WACV2023_Zhou, s_CVPR2022_ppsl, s_liu2022exploiting, s_senocak2023sound, s_song2022sspl, m_ECCV2020_Qian_coarsetofine,m_hu2020discriminative,m_hu2022mix,m_mo2023audio, m_kim2024no_prior, m_mahmud2024tvsl}. Inspired by human perception, which naturally integrates visual and auditory cues to identify the sources of sounds, this task focuses on accurately determining the spatial locations of sound-making objects within video scenes. Previous studies have explored audio-visual correspondence using self-supervised learning techniques, including hard positive mining \cite{s_hard_positive_mining}, iterative learning \cite{s_iterative2023}, and negative-free learning \cite{s_song2022sspl}. In addition, to handle more realistic and challenging settings involving multiple sound sources, generating supervisory signals \cite{m_hu2020discriminative}, graph-based modeling \cite{m_hu2022mix}, and iterative matching schemes \cite{m_kim2024no_prior} have been explored.

\begin{figure}[t]
    \begin{minipage}[b]{1.0\linewidth}
	\centering
	\centerline{\includegraphics[width=8.6cm]{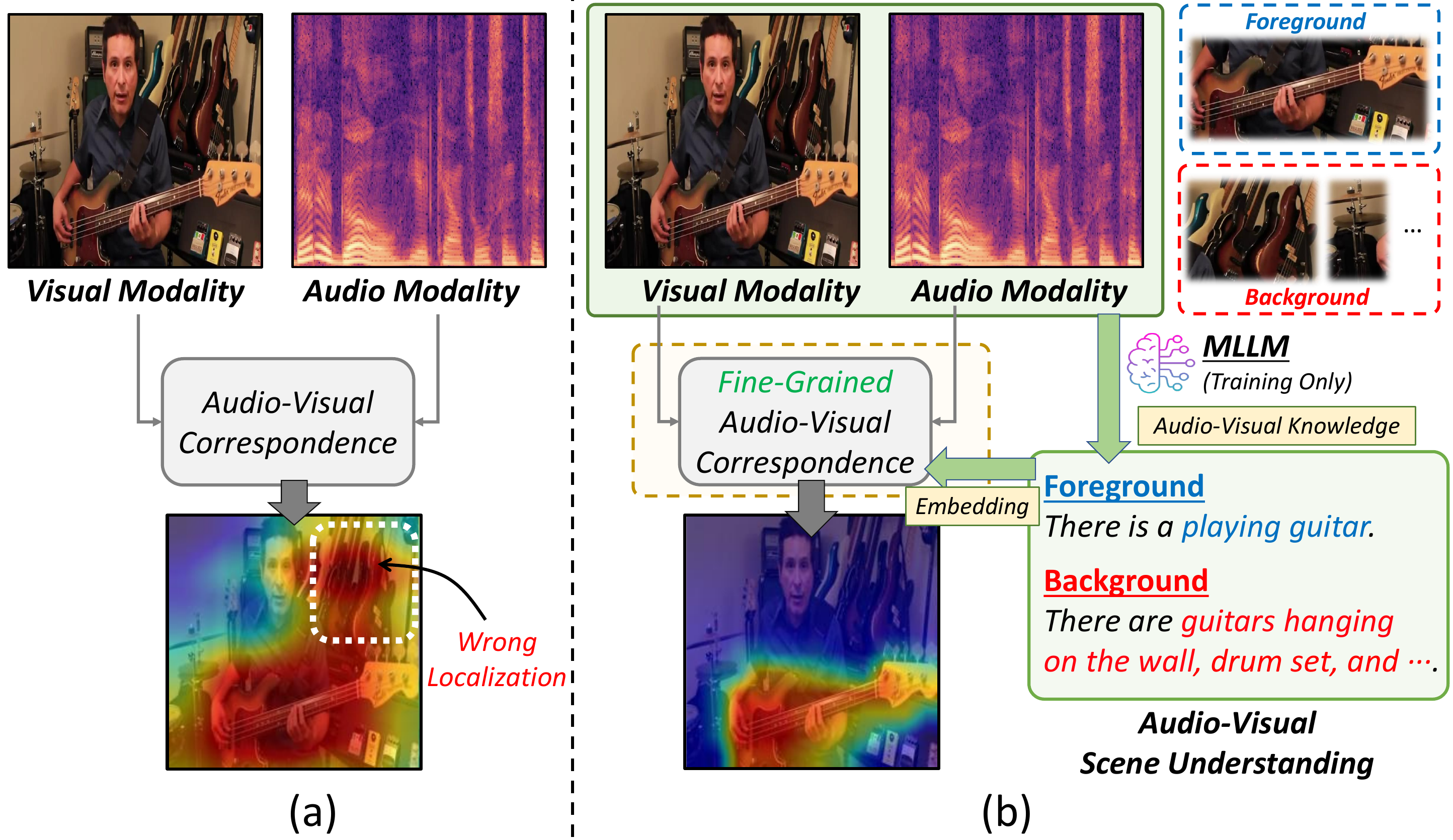}}
	\end{minipage}
        \vspace{-0.4cm}
	\caption{Conceptual comparison between (a) existing methods and (b) our method. Existing methods focus on localizing the `\textit{guitar}' object through simple audio-visual correspondence, ignoring whether it is being `\textit{played}'. In contrast, our method localizes only the playing (\textit{i.e.,} sound-making) guitar with fine-grained audio-visual correspondence from audio-visual scene understanding.}
    \vspace{-0.5cm}
    \label{fig:1}
\end{figure}




Despite these advancements, existing methods still struggle to accurately localize sound sources in complex scenes, particularly when scenes include both sound-making and visually similar silent objects simultaneously. The main limitation lies in their reliance on simple audio-visual correspondence, which does not adequately capture fine-grained differences between sound-making object and silent objects. For example, as shown in Figure \ref{fig:1}(a), existing methods mislocalize non-playing guitars by focusing solely on identifying the `\textit{guitar}' in the scene based on the sound, without considering whether the guitar is actually being `\textit{played}.' This oversimplification reduces localization accuracy, highlighting the need for a more contextually aware approach.





To address this issue, we propose a novel approach that captures more detailed contextual information by fully leveraging audio-visual scene understanding. As shown in Figure \ref{fig:1}(b), our approach achieves fine-grained localization by (\textit{i}) identifying objects present in the visual scene (\textit{e.g.,} guitars and drum set) and (\textit{ii}) distinguishing sound-making foreground objects (\textit{e.g.,} a `\textit{playing}' guitar) while differentiating silent background objects (\textit{e.g.,} `\textit{non-playing}' guitars and drum set), effectively localizing only the sound-making objects. To enable fine-grained localization through audio-visual scene understanding, we utilize Multimodal Large Language Models (MLLMs) to generate detailed guidance using their rich external knowledge during training. By incorporating MLLMs, we acquire detailed information about both sound-making and silent objects, enabling our model to learn fine-grained representations for precise audio-visual correspondence.



We further introduce two specialized loss functions during model training to effectively leverage fine-grained information. First, we propose an Object-aware Contrastive Alignment (OCA) loss that enhances the ability of our model to differentiate between sound-making and silent objects. The OCA loss leverages detailed object information from audio-visual scene understanding to align audio-visual features and accurately differentiate sound-making object from silent object, even when their visual appearances are similar. Moreover, we propose Object Region Isolation (ORI) loss to address the challenge of multiple sound sources within the same scene. This loss function promotes spatial separation of distinct sound-making objects in multi-source scenarios, enabling the model to isolate each sound source and minimize overlap between their localized regions, even for visually similar objects (\textit{e.g.,} violin and cello). By incorporating these loss functions with audio-visual scene understanding, our model effectively learns fine-grained audio-visual representations, significantly improving its ability to precisely localize and identify sound sources in complex and challenging environments.

\noindent The major contributions of our paper are as follows:
\begin{itemize}
    \item We propose a novel approach that captures more detailed contextual information by fully leveraging audio-visual scene understanding. To the best of our knowledge, this is the first attempt to address fine-grained guidance about sound-making objects and silent objects through MLLMs for audio-visual sound source localization tasks.

    \item We propose two novel loss functions, Object-aware Contrastive Alignment loss and Object Region Isolation loss, which improve fine-grained alignment and differentiation of audio-visual features, even in complex scenes.

    \item Experimental results on MUSIC and VGGSound datasets demonstrate the effectiveness of the proposed method for both single- and multi-source sound localization.
\end{itemize}
 
\section{Related Work}
\label{sec:related_work}

\subsection{Audio-Visual Sound Source Localization}
Audio-visual sound source localization aims to identify spatial positions of sound sources within video frames, typically employing self-supervised cross-modal learning methods. Early approaches primarily focus on single-source scenarios  \cite{s_CVPR2018_Senocak, s_CVPR2021_lvs, s_hard_positive_mining, s_iterative2023, s_um2023sira, s_sun2023learning, s_WACV2022_Shi, s_WACV2023_htf, s_chen2021exploring, s_momentum, s_flowgrad, s_WACV2023_Zhou, s_CVPR2022_ppsl, s_liu2022exploiting, s_senocak2023sound, s_song2022sspl}, utilizing techniques such as dual-stream attention mechanisms \cite{s_CVPR2018_Senocak}, contrastive learning targeting challenging samples \cite{s_CVPR2021_lvs, s_sun2023learning}, iterative pseudo-label refinement \cite{s_iterative2023}, optical-flow guided cross-attention \cite{s_WACV2023_htf, s_flowgrad}, enhanced semantic alignment \cite{s_senocak2023sound}, and audio-spatial integration across modalities \cite{s_um2023sira}. Recent studies expand toward more realistic multi-source environments \cite{m_ECCV2020_Qian_coarsetofine, m_hu2022mix, m_mo2023audio, m_kim2024no_prior}, proposing methods including discriminative supervisory signal generation \cite{m_hu2020discriminative}, coarse-to-fine audiovisual separation \cite{m_ECCV2020_Qian_coarsetofine}, graph-based modeling \cite{m_hu2022mix}, category-specific semantic feature extraction and integration \cite{m_mo2023audio}, iterative object identification without prior knowledge \cite{m_kim2024no_prior}, and leveraging textual-audio embeddings for complex mixtures \cite{m_mahmud2024tvsl}.
\begin{figure*}[t]
    \begin{minipage}[b]{1.0\linewidth}
	\centering
	\centerline{\includegraphics[width=16.4cm]{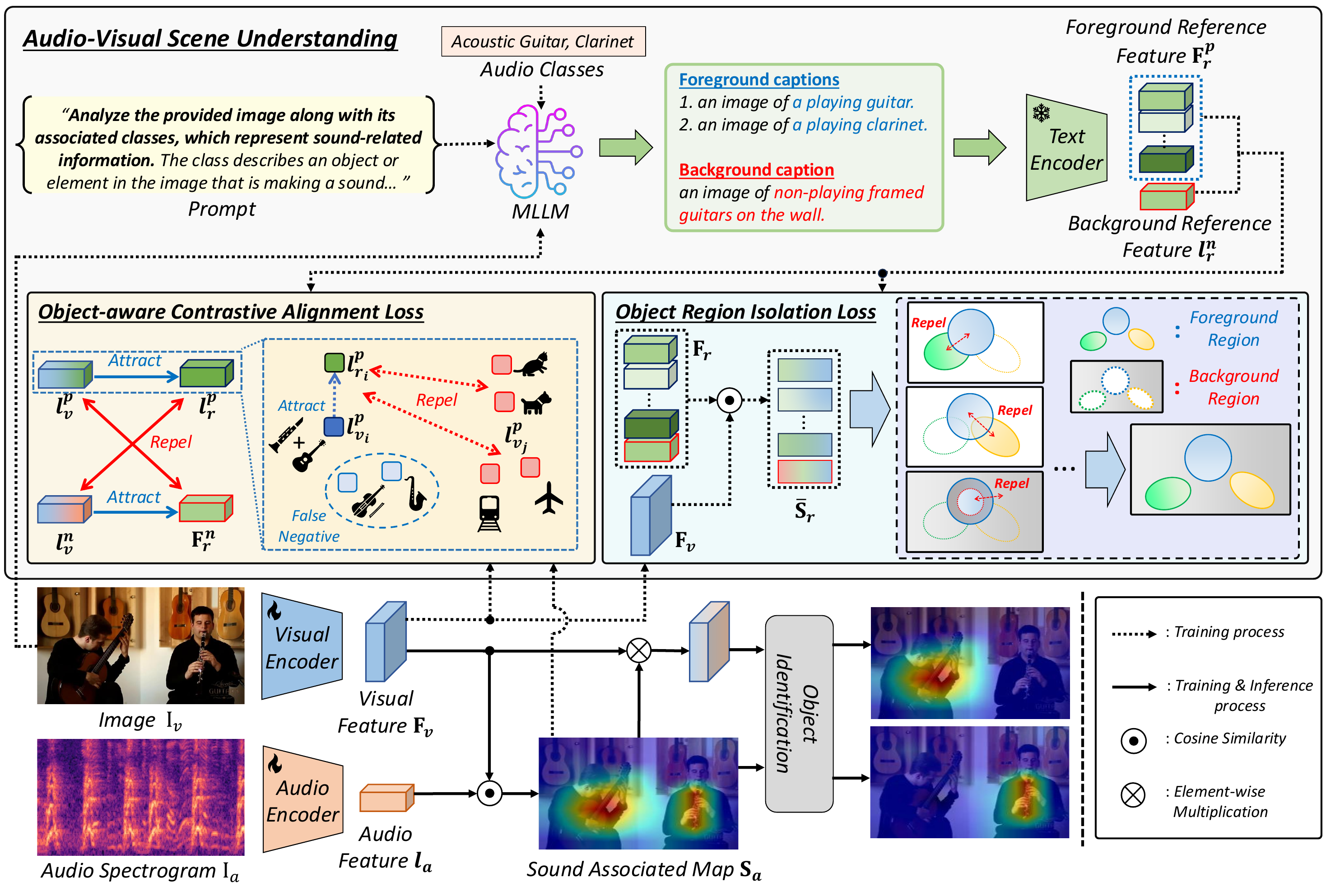}}
	\end{minipage}
    \vspace{-0.7cm}
    \caption{Network configuration of the proposed sound source localization framework. MLLM is used only during the training process and not during inference.}
    \label{fig:2} 
    \vspace{-0.4cm}
\end{figure*}


However, existing studies primarily rely on simple similarity calculations between visual and audio data, often utilizing the semantic information associated with sounds. This leads to ineffective localization when both sound-emitting and non-sound-emitting objects coexist. We propose a novel approach that captures detailed contextual information by leveraging audio-visual scene understanding, effectively bridging the audio-visual gap and enhancing fine-grained localization accuracy.


\subsection{Audio-Visual tasks with MLLMs}
Multimodal Large Language Models (MLLMs) \cite{vllm_wu2023multimodal, vllm_yin2023survey} combine the strengths of audio, visual, and language models to process complex tasks that involve both audio, image and text data. MLLMs are designed to handle various visual tasks that benefit from semantic understanding, such as object detection, scene recognition, and action detection, by integrating the contextual depth offered by language models with the spatial and visual recognition capabilities of vision models. This integration enables MLLMs to provide more interpretable outputs for audio-visual tasks.

In audio-visual tasks like audio-visual scene understanding \cite{zhang2023videollama_vllm}, video classification \cite{qian2022_videoclassification_vllm}, video sound generation \cite{wang2024v2a_audiogeneration, xie2024sonicvisionlm_audiogeneration}, audio-visual feature alignment \cite{duan2024cross_audiovisualalignment}, MLLMs enable more effective cross-modal alignment by leveraging textual descriptions to enhance the understanding of visual content associated with specific sounds. Building on this, we propose a method that leverages MLLMs for audio-visual scene understanding to generate detailed information on both sound-making and silent objects. This detailed information supports fine-grained audio-visual alignment, allowing for improved sound source localization performance.
\section{Proposed Method}
\label{sec:method}

To achieve more fine-grained sound source localization, we aim to leverage detailed contextual information that enhances audio-visual scene understanding. As shown in Figure \ref{fig:2}, during training, we leverage a comprehensive understanding of audio-visual scenes by identifying both sound-making and silent objects using Multimodal Large Language Model (MLLM) with input image and audio class information. We encode this detailed information as the reference anchor feature for fine-grained audio-visual representation learning.

We introduce two loss functions to guide fine-grained audio-visual alignment through detailed information leveraged to understand audio-visual scenes: Object-aware Contrastive Alignment (OCA) loss and Object Region Isolation (ORI) loss. These loss functions guide the model in identifying sound-making objects while ignoring silent ones and distinguishing each sound source in multi-source scenarios, ultimately improving sound source localization performance through enhanced audio-visual alignment.

\subsection{Preliminaries}
\noindent{\textbf{Sound-associated Region Localization.}} Sound source localization identifies and localizes sound-making objects in visual scenes using audio-visual information. Similar to previous works \cite{s_CVPR2018_Senocak, s_CVPR2021_lvs}, our model uses a two-stream network to extract features from both visual and audio modalities. The input image and audio spectrogram pass through each modal encoder (\textit{i.e.}, visual encoder and audio encoder) to generate visual feature $\textbf{F}_v\in \mathbb{R}^{B \times w \times h \times c}$ ($B$ indicates batch number, $w$, $h$, $c$ are the width, height, and channel) and audio feature $l_a \in \mathbb{R}^{B \times c}$, respectively. To localize sound-associated regions, we compute the cosine similarity between the visual feature $\textbf{F}_v$ and audio feature $l_a$, resulting in the sound-associated map $\textbf{S}_a$. \\

\noindent{\textbf{Object Identification.}}
For multi-sound source localization where multiple sound-making objects are present, we adopt an iterative object identification algorithm \cite{m_kim2024no_prior}. In this algorithm, object identification is performed iteratively for $K$ iterations corresponding to the number of sound sources.

\subsection{Audio-Visual Scene Understanding}
To effectively distinguish sound-making objects and silent objects in complex scenes with various objects, we utilize audio-visual understanding by using the Multimodal Large Language Model (MLLM). The MLLM, trained with extensive external knowledge of diverse and complex scenes, receives an input consisting of the given image and audio class information. Through a carefully designed prompt to distinguish whether each object is making sound or silent, we guide MLLM to effectively utilize the provided image and audio class information with some examples for complex scenes (\textit{e.g.}, if there are multiple objects that match with audio class information, distinguish between sound-making object and silent object and describe the action of objects.) This process generates $K$ foreground captions, corresponding to the number of sound sources and a single background caption. (Please refer to supplementary material for details.)


Each foreground caption provides detailed information about specific sound-making objects with their details, thereby enhancing the understanding of sound-related activities within the audio-visual scene. In contrast, background caption describes objects unrelated to sound production, offering explicit details about silent elements. By incorporating not only sound-making (foreground) but also silent (background) information, our method can understand which objects are really sound-making and silent.

The generated captions are subsequently processed through a text modality encoder to extract features, resulting in a foreground reference feature $\textbf{F}_r^p\in \mathbb{R}^{B \times K \times c}$ and a background reference feature $l_r^n\in \mathbb{R}^{B \times c}$. These features guide the model during the fine-grained audio-visual alignment process, enabling it to effectively distinguish between sound-making and non-sound-making objects, as well as differentiate between multiple sound-making objects in multi-source scenarios.



\subsection{Object-aware Contrastive Alignment Loss}
\label{sec:oca}
Existing sound source localization methods mainly rely on contrastive learning with cosine similarity for audio-visual alignment. However, since these methods focus primarily on foreground object features rather than understanding audio-visual scenes, silent objects cannot be effectively distinguished from sound-making objects. To address this issue, we propose the Object-aware Contrastive Alignment (OCA) loss. The OCA loss has an important role in learning fine-grained audio-visual correspondence through leveraging the foreground reference feature $\textbf{F}_r^p$ and background reference feature $l_r^n$ from audio-visual scene understanding.

First, we compute the cosine similarity between visual feature $\textbf{F}_v$ and audio feature $l_a$ to generate the sound-associated map $\textbf{S}_a\in\mathbb{R}^{B\times w \times h}$, defined as:

\begin{equation}
\begin{gathered}
Sim(A,B) =\frac{\langle A,B\rangle} {{||A||}\,{||B||}},\\
\textbf{S}_a = Sim(\textbf{F}_v, l_a), \\
\end{gathered}
\end{equation}
where  $\langle \cdot , \cdot \rangle$ denotes dot product. Following this, inspired by the prior work \cite{s_CVPR2021_lvs}, we create a foreground map $\textbf{M}^p$ and a background map $\textbf{M}^n$ for each foreground and background selection, defined as:
\begin{equation}
\begin{gathered}
\textbf{M}^p=\text{sigmoid}((\textbf{S}_a-\alpha_p)/\omega) \\
\textbf{M}^n=\textbf{1}-\text{sigmoid}((\textbf{S}_a-\alpha_n)/\omega), \\
\end{gathered}
\end{equation}
where $\alpha_p$, $\alpha_n$ and $\omega$ denotes the hyper-parameters \cite{s_CVPR2021_lvs}. 

Using $\textbf{M}^p$ and $\textbf{M}^n$, we then generate the foreground visual feature $l_v^p\in\mathbb{R}^{B\times c}$ and background visual feature $l_v^n\in\mathbb{R}^{B\times c}$, represented as:
\begin{equation}
\begin{gathered}
l_v^p=\text{GAP}(\langle \textbf{F}_v, \textbf{M}^p \rangle),  l_v^n=\text{GAP}(\langle \textbf{F}_v, \textbf{M}^n \rangle), \\
\end{gathered}
\end{equation}
where \text{GAP} denotes the global average pooling.

After that, we design OCA loss to enhance the ability of our model to differentiate between sound-making and silent
objects. The OCA loss is divided into a foreground alignment loss $\mathcal{L}_{frg}$ and a background alignment loss $\mathcal{L}_{bkg}$. The $\mathcal{L}_{frg}$ aims to encourage the alignment of the foreground visual feature $l_v^p$ with the average-pooled foreground reference feature $l_r^p\in\mathbb{R}^{B\times c}$, while simultaneously pushing the background visual feature $l_v^n$ away from $l_r^p$, calculated as:
\begin{equation}
\begin{gathered}
p_{i}=\exp(Sim(l_{v_i}^p, l_{r_i}^p)),\\
n_{i}^{\text{hard}} = \exp(Sim(l_{v_i}^n, l_{r_i}^p)), \\
n_i^{\text{soft}} = \sum_{j \neq i}^B\mathbbm{1}_{[ \text{Sim}(l_{r_j}^p, l_{r_i}^p) \leq \tau ]} \exp(Sim(l_{v_j}^p, l_{r_i}^p)),\\
\mathcal{L}_{frg} = -\frac{1}{B} \sum_{i=1}^{B} \log \frac{p_i}{p_i + n_{i}^{\text{hard}} + n_{i}^{\text{soft}}},
\end{gathered}
\end{equation}
where $n_i^{\text{soft}}$ handles the false negative issue \cite{huynh2022false_neg, zheng2021false_neg, dwibedi2021false_neg} inherent in batch contrastive learning, where similar sound-making objects within the batch can cause unintended negative samples. We compare average-pooled foreground reference features $l_r^p$ across different clips in the batch, applying a threshold $\tau$ to exclude samples with similarity values exceeding this threshold, as these are considered false negatives.


To fully understand the audio-visual scene, it is crucial to capture the background information through silent objects in a visual scene. To this end, the background alignment loss $\mathcal{L}_{bkg}$ aims to encourage the background visual feature $l_v^n$ to align with the background reference feature $l_r^n$, while pushing the foreground visual feature $l_v^p$ away from $l_r^n$. The $\mathcal{L}_{bkg}$ is defined as:

\begin{equation}
\begin{gathered}
\hat{p}_{i}=\exp(Sim(l_{v_i}^n, \textbf{F}_{r_i}^n)),\\
\hat{n}_{i}^{\text{hard}} = \exp(Sim(l_{v_i}^p, \textbf{F}_{r_i}^n)), \\
\mathcal{L}_{bkg} = -\frac{1}{B} \sum_{i=1}^{B} \log \frac{\hat{p}_i}{\hat{p}_i + \hat{n}_{i}^{\text{hard}}}. \\
\end{gathered}
\end{equation}
Finally, the proposed OCA loss $\mathcal{L}_{oca}$ is defined as:
\begin{equation}
\mathcal{L}_{oca} = \frac{\mathcal{L}_{frg} + \mathcal{L}_{bkg}}{2}.
\end{equation}
Through $\mathcal{L}_{oca}$, our model can learn fine-grained audio-visual representation from foreground and background object information, even in complex scenarios where both sound-making and visually similar silent objects are present.

\subsection{Object Region Isolation Loss}
\label{sec:ori}
While the audio-visual alignment achieved through the OCA loss enables effective localization of sound-making objects, multi-source scenarios require more distinct separation of sound-making objects and clear differentiation between foreground and background regions. To address this need, we propose the Object Region Isolation (ORI) loss, which enhances object separation in complex scenes by enforcing spatial distinctiveness between regions associated with each sound-making object and the background.

We generate the combined reference feature $\textbf{F}_r\in\mathbb{R}^{B\times (K+1) \times c}$ by concatenating $\textbf{F}_r^p$ and $l_r^n$. Then, using similarity calculations between the visual feature $\textbf{F}_v$ and $\textbf{F}_r$, we create a reference associated map $\textbf{S}_r=\{S_{r_k}\}_{k=1,...,K+1}\in\mathbb{R}^{B\times (K+1) \times w \times h}$, where each $S_{r_k}$ corresponds to a similarity map for each object or background region.

To further separate these regions, we employ the first-order Wasserstein Distance, also known as the Earth-Mover Distance, to measure the distance between the joint probability distributions of different objects or background regions. For this calculation, we flatten $\textbf{S}_r$ into $\bar{\textbf{S}}_r=\{\bar{S}_{r_k}\}_{k=1,...,K+1}\in\mathbb{R}^{B\times (K+1) \times wh}$. The Sinkhorn algorithm \cite{cuturi2013sinkhorn} is then applied to efficiently compute the optimal transport for minimizing the Wasserstein Distance. The loss function is defined as:
\begin{equation}
\begin{gathered}
\textbf{S}_r = Sim(\textbf{F}_v, \textbf{F}_r), \\
D_W(P, Q) = \inf_{\gamma \in \Pi(P, Q)} \mathbbm{E}_{(x,y) \sim \gamma} [\| x - y \|], \\
\mathcal{L}_{ori} = \sum_{i=1}^{B} \sum_{\substack{n,m=1 \\ n \neq m}}^{K+1}  D_W(\bar{S}_{r_n}^i, 1 - \bar{S}_{r_m}^i),\\
\end{gathered}
\end{equation}
where $D_W$ represents the first-order Wasserstein Distance.

Through $\mathcal{L}_{ori}$, the model is guided to avoid overlapping regions by encouraging clear separation in the maps generated from similarity with each reference feature. As a result, the model achieves improved separation of foreground and background regions, and in multi-source scenarios, it distinctly identifies each sound-making object within the foreground. This method ensures the model can handle complex scenes with multiple sound sources, enhancing both localization and object identification accuracy.


\begin{table*}[t]
    \renewcommand{\tabcolsep}{1.8mm}
    \centering
	\resizebox{\linewidth}{!}{
		\begin{tabular}{lccccccccc}
            \Xhline{3\arrayrulewidth}
            \rule{0pt}{10.0pt} \multirow{2}{*}[-0.5ex]{\bf Method} & \rule{0pt}{10.0pt} \multirow{2}{*}[-0.5ex]{\bf Backbone} & \multicolumn{4}{c}{\textbf{MUSIC-Duet} \cite{dataset_music}} & \multicolumn{4}{c}{\textbf{VGGSound-Duet} \cite{dataset_vggsound}} \\ 
            \cmidrule(lr){3-6} \cmidrule(l){7-10}
            && {\textbf{CAP(\%)}} & {\textbf{CloU@0.3(\%)}} & {\textbf{AUC(\%)}} & {} & {\textbf{CAP(\%)}} & {\textbf{CloU@0.3(\%)}} & {\textbf{AUC(\%)}} & {} \\
            \midrule
            Attention10k (CVPR'18) \cite{s_CVPR2018_Senocak} & VGG16 & {--} & 21.6 & 19.6 &  & {--} & 11.5 & 15.2 &  \\
            OTS (ECCV'18) \cite{s_2018ots} & VGG16 & 11.6 & 13.3 & 18.5 &  & 10.5 & 12.2 & 15.8 &  \\
            DMC (CVPR'19) \cite{s_hu2019dmc} & VGG16 & {--} & 17.5 & 21.1 &  & {--} & 13.8 & 17.1 &  \\
            CoarseToFIne (ECCV'20) \cite{m_ECCV2020_Qian_coarsetofine} & ResNet18 & {--} & 17.6 & 20.6 &  & {--} & 14.7 & 18.5 &  \\
            DSOL (NeurIPS'20) \cite{m_hu2020discriminative} & ResNet18 & {--} & 30.1 & 22.3 &  & {--} & 22.3 & 21.1 &  \\
            LVS (CVPR'21) \cite{s_CVPR2021_lvs} & ResNet18 & {--} & 22.5 & 20.9 & & {--} & 17.3 & 19.5 &  \\
            EZ-VSL (ECCV'22) \cite{s_mo2022easy} & ResNet18 & {--} & 24.3 & 21.3 &  & {--} & 20.5 & 20.2 &  \\
            Mix-and-Localize (CVPR'22) \cite{m_hu2022mix} & ResNet18 & 47.5 & 26.5 & 21.5 &  & 16.3 & 21.1 & 20.5 &  \\
            AVGN (CVPR'23) \cite{m_mo2023audio} & ResNet18 & 50.6 & 32.5 & 24.6 &  & 21.9 & 26.2 & 23.8 &  \\
            NoPrior (CVPR'24) \cite{m_kim2024no_prior} & ResNet18 & \underline{52.1} & \underline{38.6} & \underline{30.1} && \underline{32.5} & \underline{46.9} & \underline{29.2} & \\
            \cdashline{1-10}
            \rule{-2.5pt}{9.5pt}
            \bf Proposed Method & ResNet18 & \bf 61.4 & \bf 45.9 & \bf 36.1 &  & \bf 45.9 & \bf 55.2 & \bf 44.8 &  \\
            \hline  
            \rule{-2.5pt}{9.5pt}
            AudioCLIP (ICASSP'22) \cite{guzhov2022audioclip} & AudioCLIP & 56.1 & 36.9 & 29.2 & & 28.4 & 34.9 & 32.8 & \\
            AVGN (CVPR'23) \cite{m_mo2023audio} & AudioCLIP & 59.2 & 39.3 & 32.4 & & 31.9 & 37.8 & 35.4 & \\
            T-VSL (CVPR'24) \cite{m_mahmud2024tvsl} & AudioCLIP & \underline{62.9} & \underline{43.2} & \underline{35.9} & & \underline{35.7} & \underline{40.1} & \underline{37.9} & \\
            \cdashline{1-10}
            \rule{-2.5pt}{10pt}
            \bf Proposed Method & AudioCLIP & \bf 64.1 & \bf 46.9 & \bf 38.4 &  & \bf 47.1 & \bf 57.7 & \bf 47.4 &  \\
            \Xhline{3\arrayrulewidth}
            \end{tabular}
        }
    \vspace{-0.1cm}
    \caption{Experimental results on MUSIC-Duet (left) and VGGSound-Duet (right) test sets for multi-sound source localization. \textbf{Bold}/\underline{underlined} fonts indicate the best/second-best results.}
    \vspace{-0.3cm}
    \label{table:multi}
\end{table*}

\subsection{Training Objective}
To train our audio-visual scene understanding-based method, we construct the total training loss function as follows:
\begin{equation}
\begin{gathered}
\mathcal{L}_{total} = \lambda_{1}\mathcal{L}_{oca}+\lambda_{2}\mathcal{L}_{ori},
\end{gathered}
\end{equation}
where $\lambda_{1}$ and $\lambda_{2}$ denote balancing parameters for each loss function. By optimizing $\mathcal{L}_{total}$, the model learns a more robust and fine-grained audio-visual correspondence, leading to enhanced localization accuracy and better discrimination of sound sources in complex scenes.




\section{Experiment}
\label{sec:experiment}

\subsection{Datasets and Evaluation Metrics}

\noindent\textbf{MUSIC Dataset.} The MUSIC dataset \cite{dataset_music} consists of 448 unedited YouTube videos featuring solos and duets across 11 different musical instrument categories. Following prior works \cite{m_mo2023audio, m_kim2024no_prior, m_mahmud2024tvsl}, we use the same training and testing subsets to maintain consistency with previous studies. The MUSIC-Solo subset includes 358 solo videos for training and 90 solo videos for evaluation, designed for single sound source localization. Also, the MUSIC-Duet subset contains 124 duet videos for training and 17 duet videos for evaluation, aimed at multi-sound source localization tasks. \\

\noindent\textbf{VGG-Sound Dataset.} The VGG-Sound dataset \cite{dataset_vggsound}, referred to as VGGSound-Single, includes over 200k videos from 221 different sound categories. We use 144k image-audio pairs for training. For single sound source localization evaluation, we use VGG-Sound Source \cite{s_CVPR2021_lvs}. During training for multi-sound source localization, we randomly combine two video frames to generate a single input image of size $448\times224$, while also mixing the corresponding audio waveforms. Following \cite{m_mo2023audio, m_kim2024no_prior, m_mahmud2024tvsl}, we use the VGGSound-Duet dataset for evaluation. \\

\noindent\textbf{Evaluation Metrics.} For a comprehensive comparison, we adopt metrics used in prior works \cite{m_hu2022mix, m_mo2023audio, m_kim2024no_prior, m_mahmud2024tvsl}. For single sound source localization, we use Average Precision (AP), Intersection over Union (IoU), and Area Under the Curve (AUC). For multi-sound source localization, we use Class-aware Average Precision (CAP), Class-aware IoU (CIoU), and AUC.

\subsection{Implementation Details}

For the visual input, we resize images to $224\times 224$, extracting them from the center frame of each 3-second video clip. We resample the raw 3-second audio signal to 16 kHz for the audio input and convert it to a log-scale spectrogram.

Following \cite{s_CVPR2021_lvs}, we use a ResNet-18 \cite{resnet} architecture for both the visual and audio feature extraction. To handle the one-channel input of the sound spectrogram, we modify the first convolutional layer of the ResNet-18 from 3 channels to 1 channel. The visual encoder is pre-trained on ImageNet \cite{deng2009imagenet}. Additionally, for a fair comparison with T-VSL \cite{m_mahmud2024tvsl}, we also use the AudioCLIP setup, which includes pre-trained encoders with a ResNet-50 image encoder and ESResNeXt audio encoder. For the text encoder, we utilize BERT, specifically using the [CLS] token embedding from the last hidden state for sentence-level representation. We train our model using the Adam optimizer \cite{kingma2014adam} with a learning rate of $10^{-4}$ and a batch size of 128, running for 100 epochs. For MLLM, we utilize the InternVL 2.0-8B model \cite{chen2024internvl}. InternVL 2.0 is trained with an 8k context window, using training data that includes long texts, multiple images, and videos. For comparative analyses with alternative MLLMs, please refer to our supplementary materials.

\begin{table*}[t]
\small
    \renewcommand{\tabcolsep}{2.5mm}
    \centering\
	\resizebox{\linewidth}{!}{
		\begin{tabular}{lc ccc ccc}
            \Xhline{3\arrayrulewidth}
            \rule{0pt}{10.5pt} \multirow{2}{*}[-0.5ex]{\bf Method}  & \rule{0pt}{10.0pt} \multirow{2}{*}[-0.5ex]{\bf Backbone} & \multicolumn{3}{c}{\textbf{MUSIC-Solo} \cite{dataset_music}}   & \multicolumn{3}{c}{\textbf{VGGSound-Single} \cite{dataset_vggsound}} \\ 
            \cmidrule(lr){3-5} \cmidrule(l){6-8}
            && {\textbf{AP(\%)}} & {\textbf{IoU@0.5(\%)}} & {\textbf{AUC(\%)}} & {\textbf{AP(\%)}} & {\textbf{IoU@0.5(\%)}} & {\textbf{AUC(\%)}} \\
            \midrule
            Attention10k (CVPR'18) \cite{s_CVPR2018_Senocak} & VGG16 & {--} & 37.2 & 38.7  & {--} & 19.2 & 30.6  \\
            OTS (ECCV'18) \cite{s_2018ots} & VGG16 & 69.3 & 26.1 & 35.8 &   29.8 & 32.8 & 35.7   \\
            DMC (CVPR'19) \cite{s_hu2019dmc} & VGG16 & {--} & 29.1 & 38.0   & {--} & 23.9 & 27.6   \\
            CoarseToFIne (ECCV'20) \cite{m_ECCV2020_Qian_coarsetofine} & ResNet18 & 70.7 & 33.6 & 39.8   & 28.2 & 29.1 & 34.8   \\
            DSOL (NeurIPS'20) \cite{m_hu2020discriminative} & ResNet18 & {--} & 51.4 & 43.7   & {--} & 35.7 & 37.2   \\
            LVS (CVPR'21) \cite{s_CVPR2021_lvs} & ResNet18 & 70.6 & 41.9 & 40.3   & 29.6 & 34.4 & 38.2   \\
            EZ-VSL (ECCV'22) \cite{s_mo2022easy} & ResNet18 & 71.5 & 45.8 & 41.2   & 31.3 & 38.9 & 39.5   \\
            Mix-and-Localize (CVPR'22) \cite{m_hu2022mix} & ResNet18 & 68.6 & 30.5 & 40.8 &   32.5 & 36.3 & 38.9   \\
            AVGN (CVPR'23) \cite{m_mo2023audio} & ResNet18 & 77.2 & 58.1 & 48.5 &  35.3 & 40.8 & \underline{42.3}\\
            NoPrior (CVPR'24) \cite{m_kim2024no_prior} & ResNet18 & \underline{77.4} & \underline{62.1} & \underline{59.4} & \underline{46.2} & \underline{41.4} & 41.2 \\
            \cdashline{1-8}
            \rule{-2.5pt}{9.5pt}
            \bf Proposed Method & ResNet18 & \bf 79.8 & \bf 71.1 & \bf 60.9 & \bf 51.7 & \bf 47.3 & \bf 44.9   \\
            \hline
            \rule{-2.5pt}{9.5pt}
            AudioCLIP (ICASSP'22) \cite{guzhov2022audioclip} & AudioCLIP & 83.8 & 63.1 & 55.7 & 42.8 & 47.4 & 48.5 \\
            AVGN (CVPR'23) \cite{m_mo2023audio} & AudioCLIP & 85.4 & 65.8 & 56.9 & 44.1 & 49.6 & 49.5 \\
            T-VSL (CVPR'24) \cite{m_mahmud2024tvsl} & AudioCLIP & \underline{88.2} & \underline{68.5} & \underline{60.1} & \underline{48.1} & \underline{53.7} & \underline{52.9} \\
            \cdashline{1-8}
            \rule{-2.5pt}{9.5pt}
            \bf Proposed Method & AudioCLIP & \bf 89.6 & \bf 72.1 & \bf 62.5 & \bf 50.4 & \bf 57.1 & \bf 56.4  \\
            \Xhline{3\arrayrulewidth}
            \end{tabular}
        }
    \vspace{-0.2cm}
    \caption{Experimental results on MUSIC-Solo 
 (left) and VGGSound-Single (right) test sets for single sound source localization. \textbf{Bold}/\underline{underlined} fonts indicate the best/second-best results.}
    \vspace{-0.2cm}
    \label{table:single}
\end{table*}

\begin{table}[t]
    \centering
	\begin{center}
		\renewcommand{\tabcolsep}{2.1mm}
		\resizebox{0.97\linewidth}{!}
	       {
		  \begin{tabular}{cc ccc}
                \Xhline{3\arrayrulewidth}
                \rule{0pt}{10pt} $\mathcal{L}_{oca}$  & $\mathcal{L}_{ori}$ & \bf CAP(\%) & \bf CloU@0.3(\%) & \bf AUC(\%) \\ \hline
                 \rule{-2.5pt}{9.5pt}
                 -       & - & 28.8 & 31.4 & 26.2 \\
               \cmark  & - & 43.8 & 50.9 & 43.1 \\
                 -       & \cmark & 34.3 & 49.1 & 34.8 \\
                 \cdashline{1-5}
                 \rule{0pt}{10pt}
                 \cmark       & \cmark & \textbf{45.9}& \textbf{55.2} & \textbf{44.8} \\\Xhline{3\arrayrulewidth}
            \end{tabular}}
    \end{center}
    \vspace{-0.4cm}
    \caption{Effect of the two proposed losses $\mathcal{L}_{oca}$ and $\mathcal{L}_{ori}$ on VGGSound-Duet test set.}
            \label{table:loss}
       \vspace{-0.05cm}     
\end{table}

We use two RTX 4090 GPUs for training with the ResNet-18 backbone and two RTX A6000 GPUs for the AudioCLIP backbone. We set the weighting parameter as $\tau = 0.7$ for our Object-aware Contrastive Alignment (OCA) loss. For our total loss function $\mathcal{L}_{Total}$, we use $\lambda_1 = 1$ and $\lambda_2 = 0.1$. Following \cite{s_CVPR2021_lvs, m_kim2024no_prior, m_mahmud2024tvsl}, we use the provided implementation code and maintain the same hyper-parameters for other aspects of our model. 

\subsection{Comparison to Prior Works}
\noindent{\textbf{Multi-Sound Source Localization.}} We conduct experiment to compare the performance of our method with state-of-the-art multi-sound source localization methods \cite{s_CVPR2018_Senocak, s_2018ots, s_hu2019dmc, m_ECCV2020_Qian_coarsetofine, m_hu2020discriminative, s_CVPR2021_lvs, s_mo2022easy, m_hu2022mix, m_mo2023audio, m_kim2024no_prior, m_mahmud2024tvsl}. As shown in Table \ref{table:multi}, for the ResNet18 backbone, our method significantly outperforms the existing methods on the MUSIC-Duet \cite{dataset_music} test set in terms of CAP, CIoU@{0.3}, and AUC by 9.3\%, 7.3\%, and 6.0\%, respectively. In the VGGSound-Duet \cite{dataset_vggsound} test set, our approach consistently shows superior performance compared to previous methods by leveraging the audio-visual scene understanding. \\

\noindent{\textbf{Single Sound Source Localization.}} Similarly, for single sound source localization, we also conduct experiments comparing our method with existing methods \cite{s_CVPR2018_Senocak, s_2018ots, s_hu2019dmc, m_ECCV2020_Qian_coarsetofine, m_hu2020discriminative, s_CVPR2021_lvs, s_mo2022easy, m_hu2022mix, m_mo2023audio, m_kim2024no_prior, m_mahmud2024tvsl}. Table \ref{table:single} presents performances on the MUSIC-Solo \cite{dataset_music} and VGGSound-Single \cite{dataset_vggsound} datasets. Specifically, for the MUSIC-Solo test set, our method achieves improvements with the ResNet18 backbone of 2.4\% in AP, 9.0\% in IoU@{0.5}, and 1.5\% in AUC. We observe a similar tendency on the VGGSound-Single test set.

Overall, our approach consistently outperforms the existing methods across both single and multi-sound source localization tasks. This demonstrates the effectiveness of our OCA loss $\mathcal{L}_{oca}$ and ORI loss $\mathcal{L}_{ori}$ in understanding scene to enhance fine-grained audio-visual correspondence for more accurate sound source localization.

\begin{table}[t]
    \centering
    \begin{center}
        \renewcommand{\tabcolsep}{0.2mm} 
        \vspace{-0.1cm}
        \resizebox{\linewidth}{!}
        {
        \begin{tabular}{c ccc ccc}
            \Xhline{3\arrayrulewidth}
            \rule{0pt}{10pt}
            \multirow{2}{*}[-0.5ex]{\bf $\tau$} & \multicolumn{3}{c}{\bf MUSIC-Duet} & \multicolumn{3}{c}{\bf VGGSound-Duet}\\
            \cmidrule(lr){2-4} \cmidrule(l){5-7}
            & {\bf CAP(\%)} & {\bf CloU@0.3(\%)} & {\bf AUC(\%)} & {\bf CAP(\%)} & {\bf CloU@0.3(\%)} & {\bf AUC(\%)} \\ \hline
            \rule{-2.5pt}{10pt}
            0.1 & 54.1 & 41.1 & 31.2 & 36.1 & 47.2 & 31.0\\
            0.4 & 55.8 & 42.1 & 32.4 & 44.1 & 52.8 & 33.5 \\
            \bf 0.7 & \textbf{61.4} & \textbf{45.9} & \textbf{36.1} & \textbf{45.9} & \textbf{55.2} & \textbf{44.8} \\
            1.0 & 59.2 & 44.2 & 35.9 & 39.9 & 52.6 & 40.7
            \\\Xhline{3\arrayrulewidth}
        \end{tabular}}
    \end{center}
    \vspace{-0.4cm}
    \caption{Experimental results on MUSIC-Duet and VGGSound-Duet test sets according to the hyper-parameter $\tau$ for $\mathcal{L}_{oca}$.}
    \vspace{-0.2cm}
    \label{table:thr}
\end{table}



\subsection{Ablation Study}
In this section, we conduct various ablation studies to investigate (1) the effect of our proposed losses and (2) the variation of hyper-parameter $\tau$ for $\mathcal{L}_{oca}$. All ablation studies are conducted based on the VGGSound-Duet dataset with the ResNet18 backbone. \\

\noindent\textbf{Effect of the Proposed Losses.}
We first evaluate the impact of the two introduced loss functions: OCA loss $\mathcal{L}_{oca}$ and ORI loss $\mathcal{L}_{ori}$. Note that, without applying any of the proposed losses, we use the traditional audio-visual contrastive loss \cite{s_CVPR2021_lvs}. As shown in Table \ref{table:loss}, consideration of $\mathcal{L}_{oca}$ significantly improves the performance of our method with the fine-grained audio-visual correspondence compared to the baseline without this awareness. Considering $\mathcal{L}_{ori}$ also enhances the separation of overlapping regions, resulting in better localization, particularly in complex multi-source scenarios. 
We achieved the highest performance when both losses are considered, indicating that each loss uniquely contributes to improving scene understanding for sound source localization. \\

\noindent\textbf{Variation of $\tau$.} We also conduct experiments to analyze the impact of the hyper-parameter $\tau$, which controls the similarity threshold for false negatives in the OCA loss in Section \ref{sec:oca}. A lower $\tau$ value results in selecting fewer negative samples, which helps address the false negative issue by ensuring more thorough sampling of negative pairs. However, it may limit the proper understanding of similar but distinct objects, as true negatives might be misinterpreted. Conversely, a higher $\tau$ value leads to more negative samples being selected, which may reduce the effectiveness of false negative detection. Table \ref{table:thr} shows that we achieve the best performance with $\tau = 0.7$, where the most balanced training is observed.

\begin{figure}[t]
    \begin{minipage}[b]{1.0\linewidth}
	\centering
	\centerline{\includegraphics[width=6.5cm]{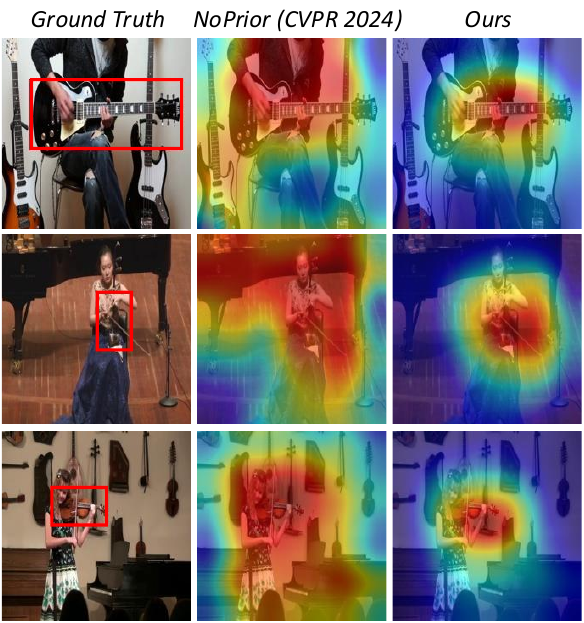}}
	\end{minipage}
        \vspace{-0.45cm}
	\caption{Visualization results for VGGSound-Single test set. We compare our method with NoPrior \cite{m_kim2024no_prior}.}
    \vspace{-0.3cm}
    \label{fig:3}
\end{figure}

\begin{figure*}[t]
    \begin{minipage}[b]{1.0\linewidth}
				\centering
				\centerline{\includegraphics[width=16.9cm]{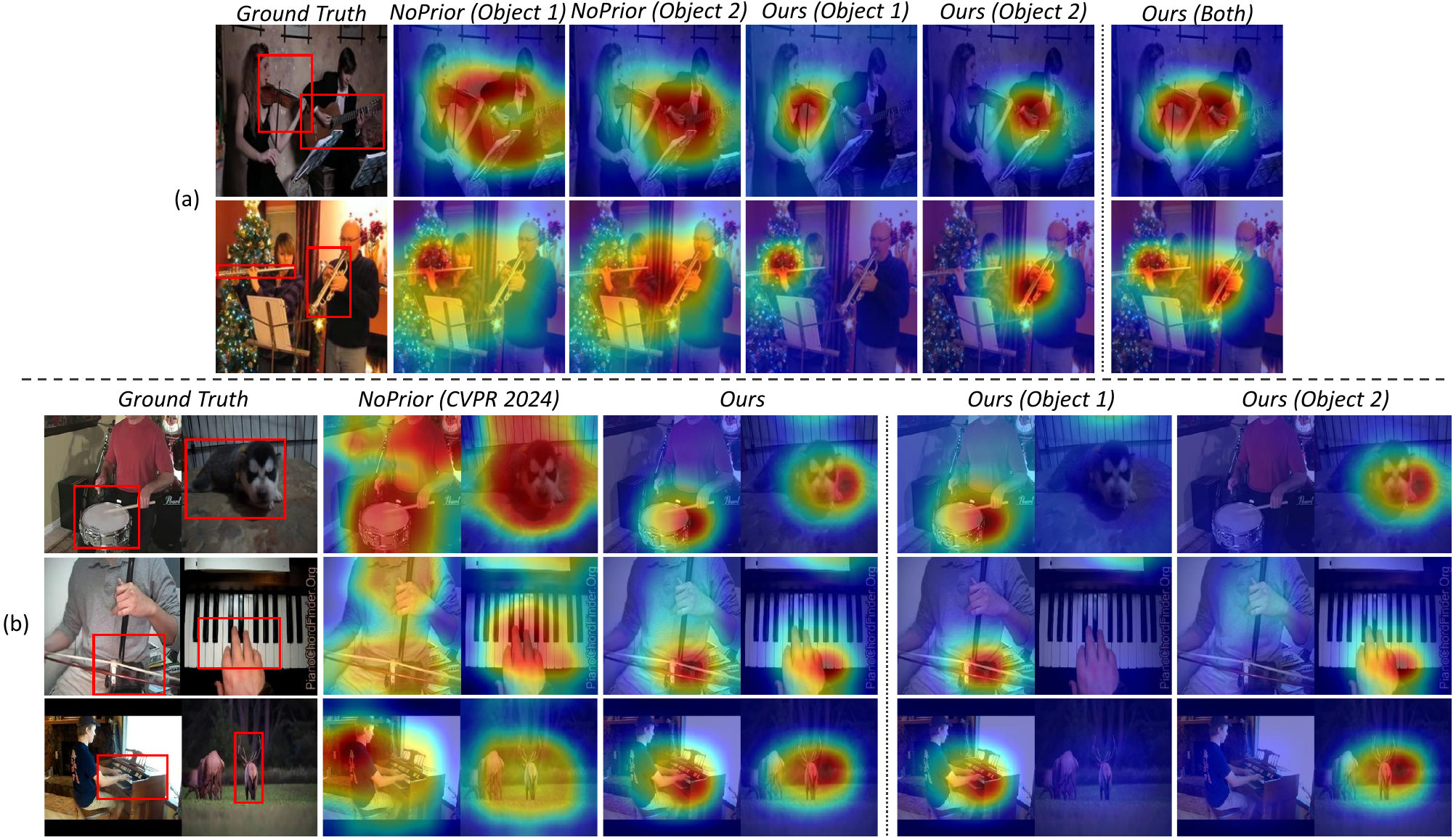}}
			\end{minipage}
			\vspace{-0.68cm}
			\caption{Visualization results for (a) MUSIC-Duet and (b) VGGSound-Duet test set. We compare our method with NoPrior \cite{m_kim2024no_prior}.}
			\vspace{-0.3cm}
   \label{fig:4}
\end{figure*}

\subsection{Visualization Results}
Figures \ref{fig:3} and \ref{fig:4} present examples of both single-sound source localization on the VGGSound-Single test set, multi-sound source localization on the MUSIC-Duet and VGGSound-Duet test set. We compare our method with the state-of-the-art method, \textit{i.e.}, NoPrior \cite{m_kim2024no_prior}. These visualization results highlight the ability of our model to perform fine-grained localization achieved with OCA loss by accurately identifying only the sound-making object. Furthermore, in multi-source scenarios, our method effectively separates sound-making objects, demonstrating the performance of the ORI loss. These results validate the effectiveness of our approach in enhancing audio-visual localization accuracy across diverse scenarios.

\begin{table}[t!]
    \centering
	\begin{center}
		\renewcommand{\tabcolsep}{1.4mm}
		\resizebox{0.97\linewidth}{!}
	    {
		  \begin{tabular}{c ccc}
                \Xhline{3\arrayrulewidth}
                \rule{0pt}{10pt} \textbf{Method} & \bf CAP(\%) & \bf CloU@0.3(\%) & \bf AUC(\%) \\ \hline 
                
                AVGN \cite{m_mo2023audio}& 18.5 & 22.7 & 21.8 \\
                
                NoPrior \cite{m_kim2024no_prior} & \underline{29.0} & \underline{34.0} & \underline{29.3} \\
                
                \cdashline{1-4}
                \rule{0pt}{9.5pt}\bf Proposed Method & \textbf{34.3} & \textbf{44.9} & \textbf{33.7}
                \\\Xhline{3\arrayrulewidth}
                \end{tabular}
            }
        \vspace{-0.2cm}
        \caption{Experimental results on VGGSound-Trio test set. \textbf{Bold}/\underline{underlined} fonts indicate the best/second-best results.}
        \label{table:trio}
    \end{center}
    \vspace{-0.75cm}
\end{table}

\subsection{Discussion}
Following previous works \cite{m_mo2023audio, m_kim2024no_prior}, we conduct additional experiments to evaluate our method with flexible number of sound sources. To this end, we use the VGGSound-Trio \cite{m_kim2024no_prior} dataset, which consists of a mixture of three sources. As shown in Table \ref{table:trio}, our method outperforms previous works across all evaluation metrics, highlighting its effectiveness in handling flexible number of sound-making objects by fully leveraging audio-visual scene understanding.
\section{Conclusion}
\label{sec:conclusion}
In this paper, we propose a novel sound source localization framework to address the challenges of sound source localization in complex scenes. To this end, our framework leverages detailed contextual information for fine-grained localization by incorporating Multimodal Large Language Models (MLLMs). By generating rich contextual details through audio-visual scene understanding, our approach enables more effective differentiation between sound-making and silent objects within various environments. Based on such contextual details, we propose two novel loss functions: Object-aware Contrastive Alignment (OCA) loss and Object Region Isolation (ORI) loss to effectively guide the model in aligning audio-visual features and ensuring distinct separation of sound sources. We believe our approach not only enhances localization accuracy but also holds diverse practical applications. \\

\noindent\textbf{Acknowledgements.} This work was supported by the NRF grant funded by the Korea government (MSIT) (No. RS-2023-00252391, RS-2025-00563942, RS-2025-00555621), and by IITP grant funded by the Korea government (MSIT) (No. RS-2022-00155911: Artificial Intelligence Convergence Innovation Human Resources Development (Kyung Hee University), RS-2022-II220124: Development of Artificial Intelligence Technology for Self-Improving Competency-Aware Learning Capabilities, IITP-2023-RS2023-00266615: Convergence Security Core Talent Training Business Support Program, No. RS-2024-00509257: Global AI Frontier Lab, IITP-2025-RS-2023-00254129: Graduate School of Metaverse Convergence Support Program)
\clearpage
{
    \bibliographystyle{ieeenat_fullname}
    \bibliography{main}
}

\clearpage

\def\paperID{698} 
\def\confName{CVPR}
\def\confYear{2025}

\def\maketitlesupplementary
   {
   \newpage
       \twocolumn[
        \centering
        \Large
        \textbf{\thetitle\\--\textit{Supplementary Material}--}\\
        \vspace{1.5em}
       ] 
   }
   
\maketitlesupplementary

\setcounter{section}{0}
\setcounter{figure}{0}
\setcounter{table}{0}
\setcounter{equation}{0}
\pagenumbering{gobble}

\noindent This manuscript provides additional implementation details and additional results of our proposed method. In Section 1, we elaborate on the additional implementation details of our method. Section 2 presents additional experimental results. Moreover, Section 3 shows additional visualization results. Note that [\textcolor{cyan}{PXX}] indicates the reference in the main paper.

\section*{1. Additional Implementation Details}

As mentioned in the main paper, we utilize the ResNet-18 [\textcolor{cyan}{P12}] for the audio encoder. At this time, since the audio spectrogram has only one channel, we modify the first convolution layer of the audio encoder to have an input channel of 1 and an output channel of 64, utilizing a kernel size of 7, stride of 2, and padding of 3. Additionally, we employ the Adam optimizer, setting the parameters $(\beta_1, \beta_2)$ to (0.9, 0.999), which are the standard values for Adam. Following [\textcolor{cyan}{P3}], we set the hyperparameters $\alpha_p=0.65$, $\alpha_n=0.4$, and $\omega=0.03$ mentioned in Section 3.3 of main paper. For our Object Region Isolation loss $\mathcal{L}_{ori}$ in Section 3.4 of main paper, we utilize the Sinkhorn algorithm [\textcolor{cyan}{P6}] with a maximum of 100 iterations. Since a reference associated map $\textbf{S}_r$ is a 2D spatial region, we incorporate both the pixel intensity differences and the Euclidean distance between the spatial coordinates of each element during the distance matrix computation.

We utilize a Multimodal Large Language Model to generate foreground captions for sound-making objects and background captions for non-sound-making objects in diverse and complex scenarios. As shown in Table \ref{prompt}, we provide prompts to guide audio-visual understanding across the following scenarios: (1) Scenarios with multiple objects, including a sound-making one, (2) Scenarios with visually similar objects, distinguishing sound-making ones, and (3) Scenarios with multiple sound-making elements. These generated foreground and background captions are employed to facilitate the learning of fine-grained audio-visual correspondence. Examples of the generated captions are provided in Section \ref{caption}.

Algorithm 1 provides additional details on the Object Region Isolation loss $\mathcal{L}_{ori}$ described in Section 3.4 in main paper. It clarifies how spatial distinctiveness is enforced between sound-making object regions and background regions by minimizing overlaps through Wasserstein Distance computation using the Sinkhorn algorithm.

\section*{2. Additional Experiments}

\noindent {\textbf{Evaluating Generalization Across Different MLLMs.}} We evaluate the generalization capability of our method by conducting experiments with various Multimodal Large Language Models (MLLMs). In addition to InternVL2.0-8B [\textcolor{cyan}{P5}], which is used in the main paper, we include two widely adopted MLLMs in recent research: LLaVA-NeXT [\textcolor{cyan}{1}] with Mistral-7B [\textcolor{cyan}{2}] and Qwen2-VL-7B [\textcolor{cyan}{3}]. These two models are also guided using the same prompt to generate foreground and background captions.

As shown in Table \ref{table:mllm}, we test the models on the VGGSound-Duet [\textcolor{cyan}{P2}] test set, where InternVL2.0-8B achieves the best performance, followed by Qwen2-VL and LLaVA-NeXT. While performance slightly varies depending on the chosen MLLM, all models consistently outperform the current state-of-the-art method, NoPrior [\textcolor{cyan}{P18}]. These results demonstrate that our approach is effective across diverse MLLMs, consistently showing superior performance in generating fine-grained audio-visual correspondences.\\

\setlength{\textfloatsep}{1pt}
\begin{algorithm}[t]
\caption{Object Region Isolation Loss Function}
\label{alg:1}
\begin{algorithmic}
\small
\Require ${\textbf{F}}_v \in \mathbb{R}^{w \times h \times c}, \textbf{F}_r \in \mathbb{R}^{(K+1) \times c}$
\Ensure $\bar{\textbf{S}}_r = Sim(\textbf{F}_v, \textbf{F}_r) \in \mathbb{R}^{wh \times (K+1)}, \mathcal{L} = 0$

\For{$i, j \in K+1, i \neq j$}
\State $\mathcal{L} = \mathcal{L} + \text{Sinkhorn}(\bar{\textbf{S}}_r[:,i], 1 - \bar{\textbf{S}}_r[:,j])$
\EndFor



\end{algorithmic}
\small
\vspace{-0.1cm}
\noindent{\textbf{Results: } {$\mathcal{L}_{ori} = \mathcal{L}$}}
\end{algorithm}
\setlength{\textfloatsep}{1pt}

\begin{table}[t]
    \centering
    \renewcommand{\tabcolsep}{0.9mm}
    \resizebox{0.97\linewidth}{!}{
        \begin{tabular}{c c ccc}
            \Xhline{3\arrayrulewidth}
            \rule{0pt}{10.5pt}
            \bf Method & \bf MLLM & \bf CAP(\%) & \bf CloU@0.3(\%) & \bf AUC(\%) \\ \hline
            \rule{0pt}{10.5pt}
            NoPrior [\textcolor{cyan}{P18}] & {--} & 32.5 & 46.9 & 29.2 \\ 
            \cdashline{1-5}
            \rule{0pt}{10pt}
            \multirow{3}{*}{\rule{0pt}{9.5pt}\shortstack{\textbf{Proposed} \\ \textbf{Method}}} & LLaVa-NeXT [\textcolor{cyan}{1}] & 41.4 & 49.8 & 42.2 \\
            & Qwen2-VL [\textcolor{cyan}{3}] & 43.1 & 54.4 & 43.9 \\
            & InternVL2 [\textcolor{cyan}{P5}] & \textbf{45.9} & \textbf{55.2} & \textbf{44.8} \\ 
            \Xhline{3\arrayrulewidth}
        \end{tabular}
    }
    \caption{Experimental results on the VGGSound-Duet test set using different Multimodal Large Language Models (MLLMs).}
    \label{table:mllm}
\end{table}


\begin{table}[t]
    \centering
    \renewcommand{\tabcolsep}{1.0mm}
    \resizebox{\linewidth}{!}{
        \begin{tabular}{cc c cc}
            \Xhline{3\arrayrulewidth}
            \rule{0pt}{10.5pt}
            \bf Method & \bf Text Encoder & \bf CAP(\%) & \bf CloU@0.3(\%) & \bf AUC(\%) \\ 
            \hline
            NoPrior [\textcolor{cyan}{P18}] & {--} & 32.5 & 46.9 & 29.2 \\ 
            \cdashline{1-5}
            \rule{0pt}{10pt}
            \multirow{2}{*}{\rule{0pt}{9.5pt}\shortstack{\textbf{Proposed} \\ \textbf{Method}}} & \rule{0pt}{9.5pt}
            CLIP [\textcolor{cyan}{4}] & 43.7 & 54.0 & 42.7 \\
            & BERT [\textcolor{cyan}{5}] & \textbf{45.9} & \textbf{55.2} & \textbf{44.8} \\
            \Xhline{3\arrayrulewidth}
        \end{tabular}
    }
    \caption{Experimental results on the VGGSound-Duet test set using different text encoders for generated caption.}
    \label{table:text_encoder}
\end{table}


\begin{figure*}[t]
    \begin{minipage}[b]{1.0\linewidth}
				\centering
				\centerline{\includegraphics[width=16cm]{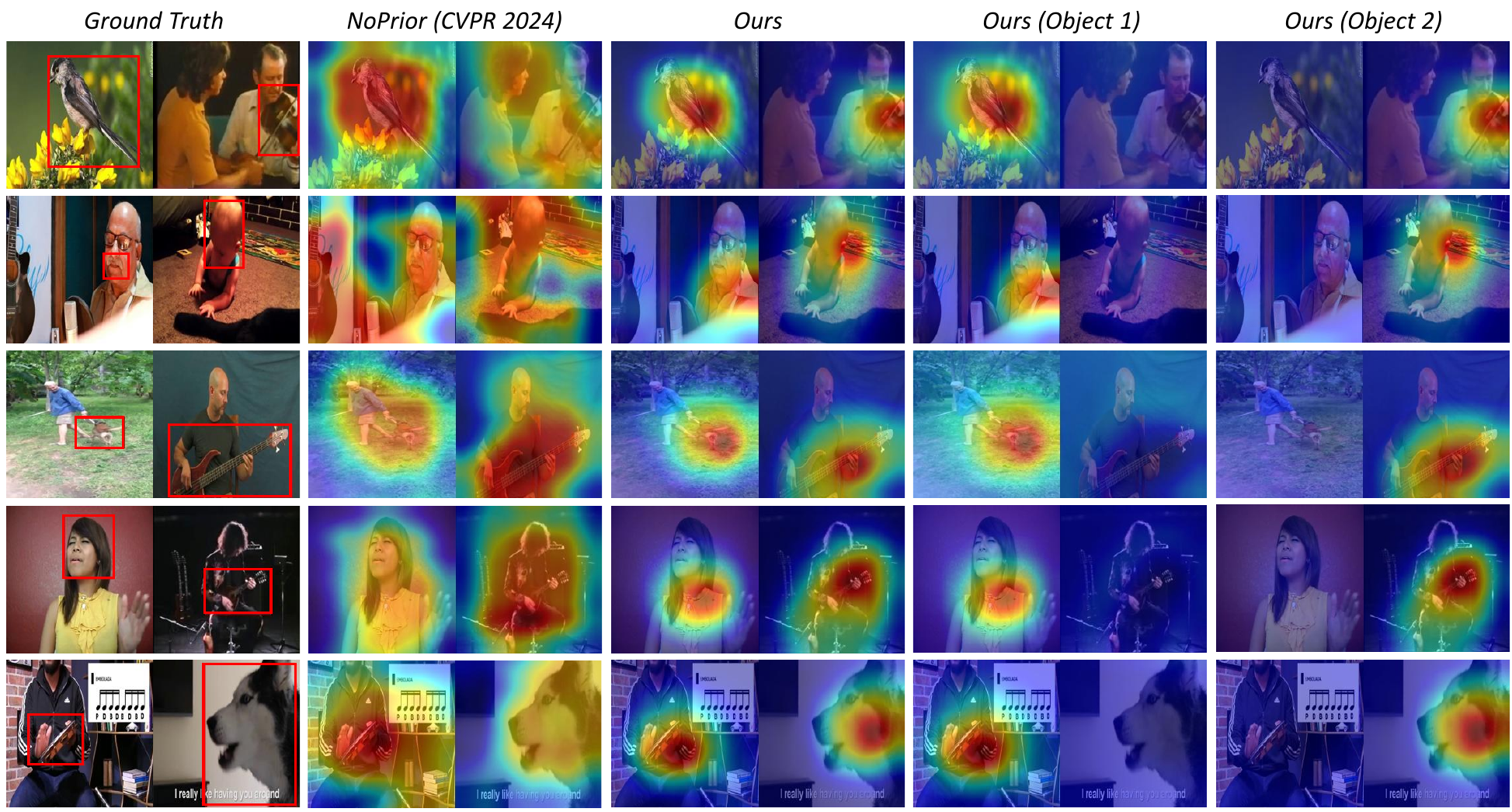}}
			\end{minipage}
			\vspace{-0.49cm}
			\caption{Additional visualization results for VGGSound-Duet test set.}
            \vspace{-0.1cm}
   \label{fig:vis}
\end{figure*}

\noindent {\textbf{Performance Variation with Different Text Encoders.}}
We present an additional experiment to validate the robustness of our approach with various text encoders for generated captions in Section 3.2 of the main paper. While the main paper used BERT as the default text encoder, we additionally adopt CLIP [\textcolor{cyan}{4}], widely used for image-text multimodal learning, to further investigate the generalization ability of our method across different text encoders. We utilize the VGGSound-Duet test set to compare performance. As shown in Table \ref{table:text_encoder}, our method significantly outperforms the existing approaches across both tested text encoders. These results indicate the effectiveness and generalization ability of our model in incorporating different text encoder models. \\\\\\


\begin{table}[t]
    \centering
    \begin{center}
		\renewcommand{\tabcolsep}{2.2mm}
		\resizebox{0.99\linewidth}{!}
	       {
		  \begin{tabular}{cc cc cc}
                \Xhline{3\arrayrulewidth}
                \rule{0pt}{9.5pt}
                \bf Method & $\lambda_1$ & $\lambda_2$ & \bf CAP(\%) &\bf CloU@0.3(\%) & \bf AUC(\%) \\ \hline
                \rule{0pt}{9.5pt}
                NoPrior [\textcolor{cyan}{P18}] & {--} & {--} & 32.5 & 46.9 & 29.2 \\ 
                \cdashline{1-6}
                \rule{0pt}{9.5pt}
                \multirow{8}{*}{\rule{0pt}{9.5pt}\shortstack{\textbf{Proposed} \\ \textbf{Method}}} & \multirow{4}{*}{\makecell{\textbf{1}}} & 10 & 42.2 & 51.1 & 40.6 \\
                
                && 1 & 45.4 & 53.5 & 44.4 \\
                
                && \textbf{0.1} & \bf 45.9 & \bf 55.2 & \bf 44.8 \\
                
                && 0.01 & 45.2 & 51.5 & 42.2 \\
                
                 \cdashline{2-6}
                \rule{0pt}{9.5pt}
                & 10 & \multirow{4}{*}{\makecell{\textbf{0.1}}} & 43.1 & 53.3 & 42.9 \\
                
                & \textbf{1} & & \bf 45.9 & \bf 55.2 & \bf 44.8 \\
                
                & 0.1 &  & 43.2 & 51.3 & 42.1 \\
                
                & 0.01 &  & 42.6 & 50.1 & 41.2
                 \\\Xhline{3\arrayrulewidth}
                \end{tabular}}
    \end{center}
    \vspace{-0.5cm}
    \caption{Experimental results on the VGGSound-Duet test set using different balancing parameters $\lambda_1$ and $\lambda_2$.}
    \label{table:balancing_parm}
\end{table}

\noindent {\textbf{Experiment on Balance Parameters $\lambda_1$ and $\lambda_2$ used in Loss Function.}}
To evaluate the effect of the balancing parameters $\lambda_1$ and $\lambda_2$ in our total loss function in Section 3.5 of main paper, we perform an additional study. As shown in Table \ref{table:balancing_parm}, our model achieves optimal performance when $\lambda_1 = 1$ and $\lambda_2 = 0.1$. Remarkably, our method consistently surpasses the existing approach across a range of balancing parameters. These results demonstrate that even when varying the hyperparameters, our method maintains consistently high performance with minimal variation, indicating that our approach does not heavily rely on hyperparameter tuning. \\

\noindent {\textbf{Computational Cost.}}
We compare the computational efficiency of our method with T-VSL [\textcolor{cyan}{P23}], which uses AudioCLIP as its backbone. As shown in Table \ref{table:cost}, despite incorporating MLLMs during training, our approach maintains comparable efficiency to state-of-the-art methods. For the MUSIC dataset (50K samples, 100 epochs), T-VSL requires 47.8h for training, while our method takes 48.1h. Importantly, MLLMs are only used at the beginning of training for caption generation and not during inference, ensuring real-world applicability. Our method demonstrates improved efficiency with 13.7\% faster inference time and lower memory usage compared to T-VSL [\textcolor{cyan}{P23}], due to our simpler architecture and the absence of MLLMs at test time.
\\



		\begin{table}[t]
			\renewcommand{\tabcolsep}{0.7mm}
			\centering
			\resizebox{0.96\linewidth}{!}{
				\begin{tabular}{c c cc cc}
					\Xhline{3\arrayrulewidth}
					\rule{0pt}{9.5pt} \multirow{3}{*}{\bf Method} & \multirow{2}{*}{\bf MLLM } & \multicolumn{2}{c}{\bf Training} & \multicolumn{2}{c}{\bf Inference}\\
                    \cmidrule(r){3-4}\cmidrule(r){5-6}&\multirow{2}{*}{\textbf{time} (s)}&
					\multirow{2}{*}{\textbf{time} (s)}&\multirow{2}{*}{\textbf{memory} (GB)}&\multirow{2}{*}{\textbf{time} (s)}&\multirow{2}{*}{\textbf{memory} (GB)}\\
                    \rule{0pt}{10pt}&\textit{(per image)}&\textit{(per iter)} & &\textit{(per image)} & 					\\ \hline
		
            \rule{0pt}{9.5pt}
                    T-VSL [\textcolor{cyan}{P23}] & - & 1.53 & 22.0 & 0.051 & 0.81 \\
					\textbf{Ours} & 0.92 & 1.13 & 16.63 & 0.044 & 0.35 \\
					\Xhline{3\arrayrulewidth}
				\end{tabular}
			}
            \vspace{-0.3cm}
            \caption{The comparisons of training time, inference time, and the number of parameters.}

			\label{table:cost}
		\end{table}

\section*{3. Additional Visualization Results}
\noindent {\textbf{Visualization Results on VGGSound-Duet.}} We provide additional visualization results in Figure \ref{fig:vis} to demonstrate the effectiveness of our method in achieving fine-grained audio-visual localization. Our approach identifies sound-making objects, leveraging the Object-aware Contrastive Alignment loss $\mathcal{L}_{oca}$ to focus exclusively on sound-making regions. Building on this, the Object Region Isolation loss $\mathcal{L}_{ori}$ enhances the ability of model to separate multiple sound-making objects in multi-source scenarios, ensuring precise isolation of each source. Together, these two loss functions enable the model to handle diverse audio-visual scenes effectively. These results highlight the effectiveness of our approach in improving audio-visual localization accuracy across a wide range of challenging scenarios.\\


\begin{table*}[htbp]
    \centering
    \begin{mybox2}

\textbf{Analyze the provided image along with its associated class label, which identifies an object or element in the image that emits sound. The scene is complex, containing multiple objects, and requiring categorization based on the examples below.}\\

\textit{Instructions}:  \\
1. Identify foreground (sound-related) elements: These are objects in the image emitting sounds that match the class description.\\
2. Identify background (sound-unrelated) elements: These are distinct objects visible in the image but unrelated to the sound described by the class.\\
3. Focus strictly on what is visible in the image. Do not infer or describe unseen objects.\\

\textit{Output Format}: \\
The response must always be in JSON format with structured sentences that start with `an image of....'. If there are two or more class labels (separated by commas), the foreground must be provided as a list of sound-making elements.\\

\textit{Examples}:\\
\textbf{(1) Scenario with multiple objects, including a sound-making one}\\
Input:\\
- image: $\text{example\_image\_1}$\\
- class label: $\text{man\_blowing\_whistle}$

Output: 
\begin{verbatim}
{
    "foreground": "an image of a man blowing a whistle",
    "background": "an image of mountains, desert landscape, and sky"
}
\end{verbatim}

\textbf{(2) Scenario with visually similar objects, distinguishing sound-making ones}\\
Input:\\
- image: $\text{example\_image\_2}$\\
- class label: $\text{acoustic\_guitar}$

Output: 
\begin{verbatim}
{
    "foreground": "an image of a man playing guitar",
    "background": "an image of non-playing guitars, drum-set, and amp"
}
\end{verbatim}

\textbf{(3) Scenario with multiple sound-making elements}\\
Input:\\
- image: $\text{example\_image\_3}$\\
- class label: $\text{clarinet, violin}$

Output: 
\begin{verbatim}
{
    "foreground": ["an image of playing clarinet", "an image of playing violin"],
    "background": "an image of the kitchen, curtains, and piano in the background"
}
\end{verbatim}
Now, process the provided input following the same structure and RETURN ONLY the JSON FORMAT.
    \end{mybox2}
     \caption{Guiding prompt for foreground and background caption generation using a Multimodal Large Language Model (MLLM)}
     \label{prompt}
\end{table*}

\begin{figure*}[t]
    \begin{minipage}[b]{1.0\linewidth}
				\centering
                \centerline{\includegraphics[width=18cm]{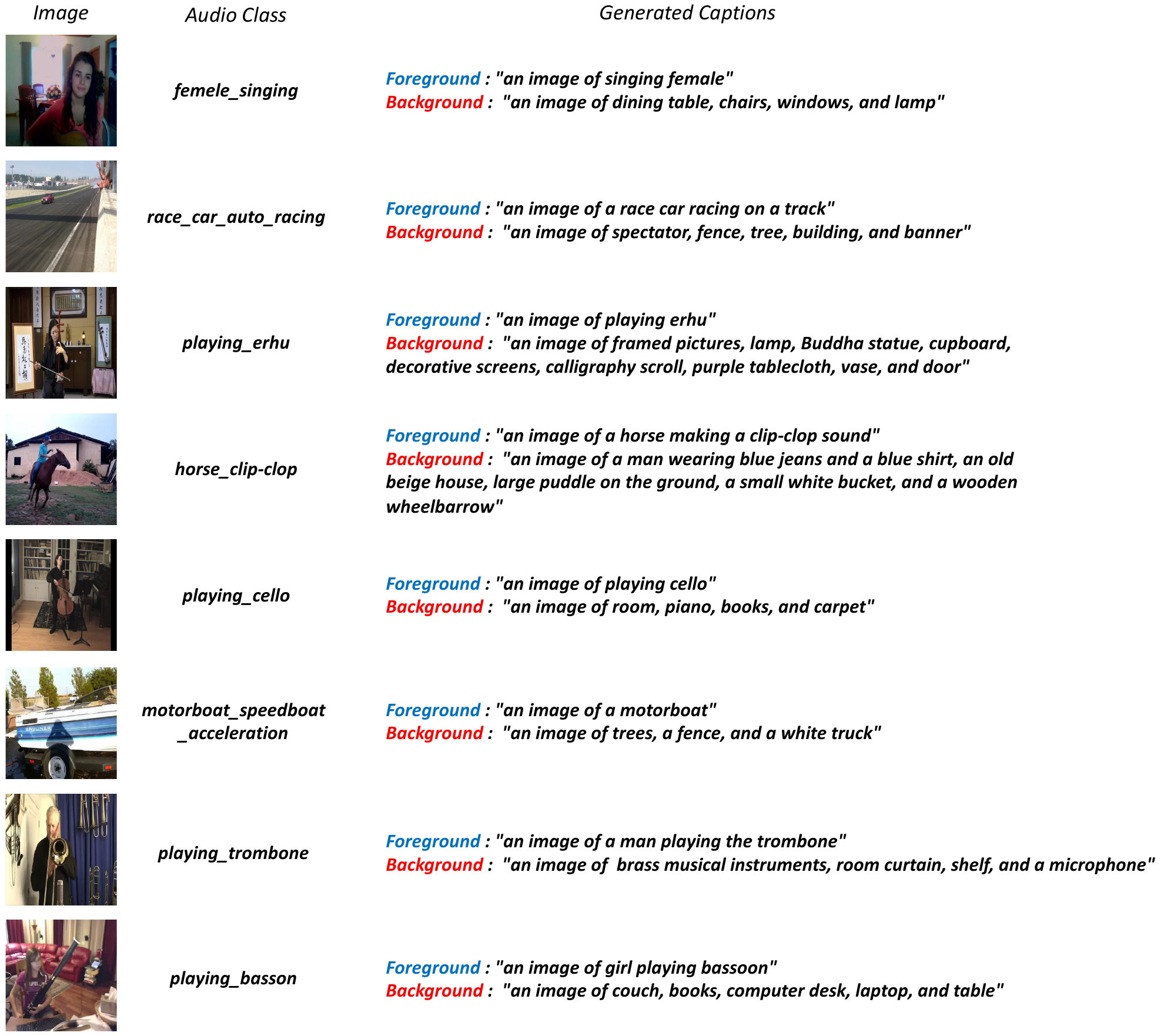}}
			\end{minipage}
			\vspace{-0.69cm}
			\caption{Visualization results of generated captions on VGGSound train set.}
   \vspace{-0.1cm}
   \label{fig:cap}
\end{figure*}

\noindent {\label{caption}\textbf{Quality of the Generated Captions.}} We visualize the generated foreground and background captions, along with the corresponding image and audio class information. The captions are generated using the InternVL 2.0-8B model, guided by a carefully crafted prompt (refer to Table \ref{prompt}). Foreground captions describe the sound-making objects corresponding to the provided audio class, while background captions capture the silent objects visible in the image. Figure \ref{fig:cap} demonstrates the consistency and relevance of the captions in relation to the audio-visual scenes, highlighting how effectively MLLMs capture both sound-making and silent objects in diverse scenarios. \\

\noindent {\textbf{Video Demo.}} We provide supplementary video materials that offer a more in-depth explanation of our method for localizing sound-making objects in complex environments. These videos demonstrate the real-time applicability and robustness of our approach under various conditions. We provide the results of our method with some examples from the VGGSound-Single and VGGSound-Duet datasets, illustrating its ability to distinguish sound-making objects from silent ones effectively. Please refer to the video titled ``CVPR2025\_SubmissionID\_698\_Supp\_Demo.mp4".


\section*{References}
\small [1] Liu, Haotian, et al. Visual instruction tuning. In \textit{NeurIPS}, 2024. \\
\small [2] Jiang, Albert Q., et al. Mistral 7B. In \textit{arXiv preprint}, 2023. \\
\small [3] Wang, Peng, et al. Qwen2-vl: Enhancing vision-language model's perception of the world at any resolution. In \textit{arXiv preprint}, 2024. \\
\small [4] Radford, Alec, et al. Learning transferable visual models from natural language supervision. In \textit{ICML}. PMLR, 2021. \\
\small [5] Devlin, Jacob, et al. BERT: Pre-training of Deep Bidirectional Transformers for Language Understanding. In \textit{arXiv preprint}, 2019. \\
\small [P2] Honglie Chen, et al. Vggsound: A large-scale audio-visual dataset. In \textit{ICASSP}, 2020. \\
\small [P3] Honglie Chen, et al. Localizing visual sounds the hard way. In \textit{CVPR}, 2021. \\
\small [P5] Zhe Chen, et al. Internvl: Scaling up vision foundation models and aligning for generic visual-linguistic tasks. In \textit{CVPR}, 2024.\\
\small [P6] Marco Cuturi. Sinkhorn distances: Lightspeed computation of optimal transport. In \textit{NeurIPS}, 2013. \\
\small [P12] Kaiming He, et al. Deep residual learning for image recognition. In \textit{CVPR}, 2016. \\
\small [P18] Dongjin Kim, et al. Learning to visually localize sound sources from mixtures without prior source knowledge. In \textit{CVPR}, 2024. \\
\small [P23] Tanvir Mahmud, et al. T-vsl: Text-guided visual sound source localization in mixtures. In \textit{CVPR}, 2024. \\


\end{document}